\documentclass[10pt,twocolumn,letterpaper]{article}

\usepackage[pagenumbers]{cvpr} 

\usepackage[dvipsnames]{xcolor}

\usepackage{graphicx}
\usepackage{amsmath}
\usepackage{amssymb}
\usepackage{amsfonts}
\usepackage{adjustbox}
\usepackage{booktabs}
\usepackage{multirow}
\usepackage{multicol}
\usepackage{float}
\usepackage{mathtools}
\usepackage{pifont}
\usepackage[accsupp]{axessibility}

\definecolor{cvprblue}{rgb}{0.21,0.49,0.74}
\usepackage[pagebackref,breaklinks,colorlinks,citecolor=cvprblue]{hyperref}

\def\paperID{1} 
\def\confName{RHOBIN}
\def\confYear{2024}

\title{MoCap-to-Visual Domain Adaptation for Efficient Human Mesh Estimation from 2D Keypoints}

\author{Bedirhan Uguz$^1$, Ozhan Suat$^1$, Batuhan Karagoz$^1$, Emre Akbas$^{1,2}$ \\
$^1$Department of Computer Engineering and $^2$METU ROMER Robotics Center\\
Middle East Technical University, Ankara, Turkey\\
{\tt\small \{bedirhan.uguz, ozhan.suat, batuhan.karagoz, eakbas\}@metu.edu.tr}}

\begin{document}
\maketitle
\begin{abstract}
This paper presents Key2Mesh, a model that takes a set of 2D human pose keypoints as input and estimates the corresponding body mesh. Since this process does not involve any visual (i.e. RGB image) data, the model can be trained on large-scale motion capture (MoCap) datasets, thereby overcoming the scarcity of image datasets with 3D labels. To enable the model's application on RGB images, we first run an off-the-shelf 2D pose estimator to obtain the 2D keypoints, and then feed these 2D keypoints to Key2Mesh. 
To improve the performance of our model on RGB images, we apply an adversarial domain adaptation (DA) method to bridge the gap between the MoCap and visual domains. 
Crucially, our DA method does not require 3D labels for visual data, which enables adaptation to target sets without the need for costly labels. We evaluate Key2Mesh for the task of estimating 3D human meshes from 2D keypoints, in the absence of RGB and mesh label pairs. Our results on widely used  H3.6M and 3DPW datasets show that Key2Mesh sets the new state-of-the-art by outperforming other models in PA-MPJPE for both datasets, and in MPJPE and PVE for the 3DPW dataset.
Thanks to our model's simple architecture, it operates at least 12$\times$ faster than the prior state-of-the-art model, LGD \cite{song2020human:LGD}. Additional qualitative samples and code are available on the project website: \url{https://key2mesh.github.io/}.
\end{abstract}    
\section{Introduction}
\label{sec:intro}

Accurate estimation of 3D human pose and shape (HPS) from a single-view image is a challenging problem in computer vision with many applications in human motion understanding \cite{zhang2021learning} and generation \cite{petrovich2021action}, AR/VR \cite{weng2019photo}, human-computer interaction \cite{guzov2021human}, and medical field \cite{grimm2012markerless, chen2018patient, rodrigues2022multi}. Parametric human body models like SMPL \cite{SMPL:2015} have played a pivotal role in driving recent advancements in this domain. A typical HPS estimation method \cite{kanazawa2018end:HMR, kocabas2019vibe, li2021hybrik, lin2021end-to-end, cho2022FastMETRO,  ma20233d:VirtualMarkers,  tripathi20233d} takes an image (or video) as input and estimates SMPL parameters. Mostly, such methods follow a fully supervised approach and rely on the availability of image datasets with 3D annotations. However, obtaining such ``paired'' annotations is both costly and challenging, particularly in the wild. One usually needs to set up an expensive motion capture system that is extensively tailored for the environment. 
Consequently, existing datasets for direct 3D supervision are small and 
do not cover the complete range of variations in human body (shape, size, pose, appearance), and scene (lightning, environment). In addition to technical challenges, the practical application of traditional data collection methods faces significant challenges in acquiring comprehensive visual data, particularly in sensitive fields like the medical domain, where privacy concerns restrict full-image collection.

Several prior work have addressed the shortage of 3D labels by generating pseudo-labels through multiview geometry \cite{kocabas2019epipolar}, optimization \cite{kolotouros2019spin}, and fine-tuning a model \cite{joo2020eft, li2022cliff}. Another set of approaches explores the usage of 2D labels as auxiliary information sources. The availability of large image and video datasets made it possible to reliably estimate and train image-based HPS models with 2D cues such as 2D keypoints \cite{kanazawa2018end:HMR, kolotouros2019spin, kocabas2019vibe, lin2021end-to-end, choi2020beyond, tripathi2020posenet3d, li2022cliff, tripathi20233d, cho2023implicit}, silhouettes \cite{pavlakos2018learning, yu2021skeleton2mesh}, optical flow \cite{tung2017self}, and body part segmentations \cite{zanfir2020weakly, zanfir2021neural}.
Unlike 2D annotations, large-scale 
MoCap datasets like AMASS \cite{AMASS:ICCV:2019} provide extensive 3D details, spanning body pose, shape distributions, and mesh vertex configurations. However, incorporating such a dataset in HPS estimation is not trivial since it does not contain any RGB data (hence, ``unpaired’’). 

Some image-based methods leverage unpaired MoCap data to supervise their models with adversarial losses \cite{kanazawa2018end:HMR, kocabas2019vibe} or establish human body priors \cite{kolotouros2021probabilistic, rempe2021humor}. Moreover, as a further step, recent work employs various proxy representations such as 2D keypoints \cite{song2020human:LGD, choutas2022learning:FMM}, silhouettes \cite{sengupta2020synthetic:STRAP, sengupta2021probabilistic:STRAPV2}, edge images \cite{sengupta2021hierarchical:STRAPV3} and dense correspondences \cite{gong2022cra} to train models with unpaired MoCap data. 
Common to these methods is the generation of ``proxy representation (e.g. 2D keypoints)"-``human mesh" training pairs from MoCap samples. 
During inference, obtaining proxy representations is straightforward through readily available estimators trained on the aforementioned large 2D datasets. Also, using proxy representations offers a strategic workaround for data collection challenges in sensitive fields by avoiding the need to store identifiable image data, thus addressing privacy concerns. However, as this process only involves MoCap data, it falls short of accurately capturing visual data attributes, disregarding crucial factors such as estimation errors, occlusions, and noise introduced by the estimators. 

\begin{figure}

\centering 

\includegraphics[width=0.47\textwidth]{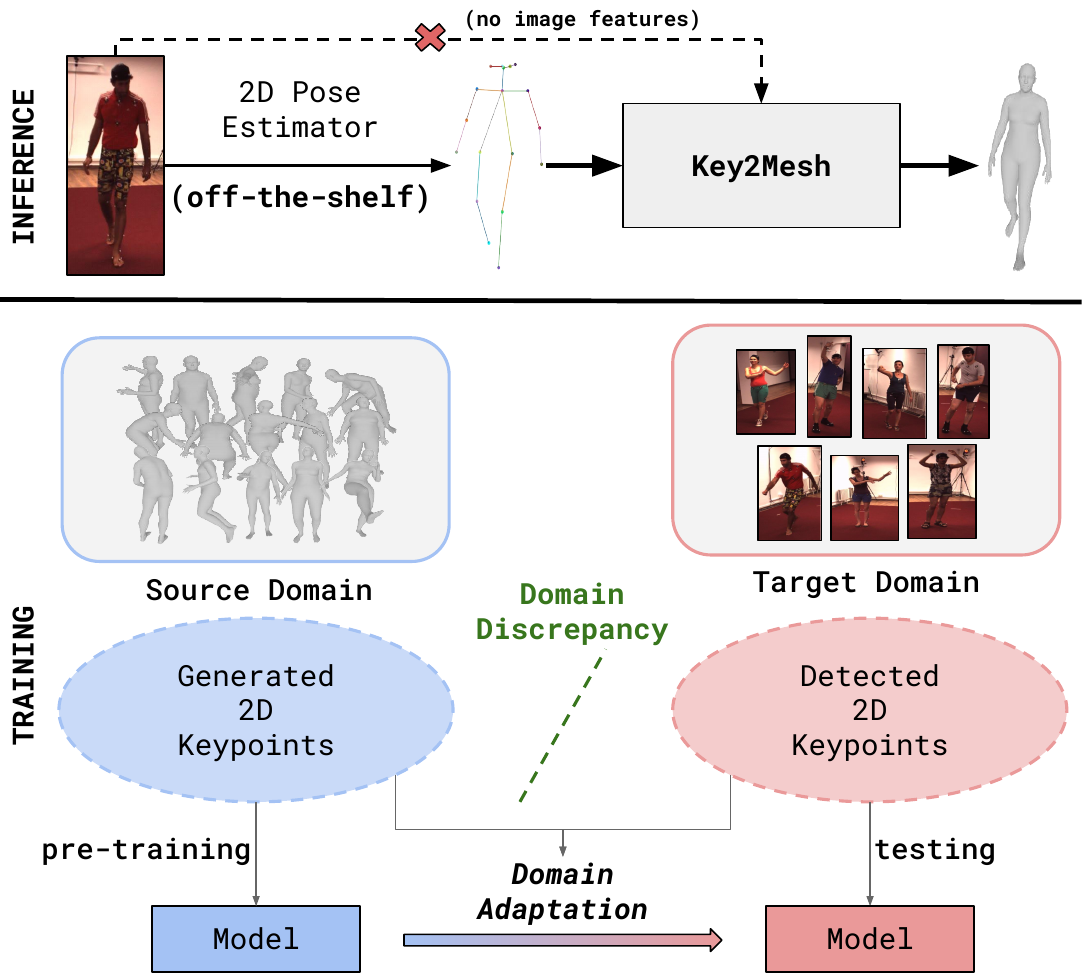}

\caption{\textbf{An overview of Key2Mesh's inference and training process.} (Top) \textbf{Inference:} For a given input image, we run an off-the-shelf 2D  pose estimator to obtain 2D keypoints, which are input to our Key2Mesh model to estimate a body mesh.
(Bottom) \textbf{Training:} Source domain consists of 3D body meshes obtained from MoCap data. We generate 2D keypoints from these body meshes using virtual cameras and a range of augmentations. We then pre-train our model using these generated 2D keypoints and corresponding meshes, without incorporating any RGB images. Finally, we adapt our model to target RGB images by using an off-the-shelf 2D pose estimator and bridging the gap between the ``2D keypoints obtained from MoCap’’ domain and the ``2D keypoints obtained from the target RGB images’’ domain. }

\label{fig:teaser}

\end{figure}

In this paper, we introduce Key2Mesh, a model estimating a human body mesh from 2D keypoints, applied to RGB images after extracting keypoints via an off-the-shelf 2D human pose estimator (\cref{fig:teaser} inference). We pre-train Key2Mesh on the body mesh instances obtained from a large-scale unpaired MoCap dataset. Specifically, we extract 2D keypoints from a body mesh in the dataset using a random virtual camera, which serves as input for our model. Then, Key2Mesh is pre-trained on such 2D keypoint - body mesh pairs (\cref{fig:teaser}  training).

To bridge the gap between the source (MoCap) and target (Visual) domains, we introduce a domain adaptation process that utilizes detected 2D keypoints from the target domain and further train our Key2Mesh model. We employ a domain critic and train the Key2Mesh model in an adversarial manner similar to ADDA \cite{tzeng2017adversarial:ADDA}, along with a re-projection loss and a feature regularization to support HPS estimation task on the target domain (\cref{fig:teaser} training). Notably, our method does not require 3D labels from the target domain, making it easy to adapt the pre-trained Key2Mesh model from the source domain to different target domains.

We evaluate our approach on two widely-used indoor and outdoor 3D human pose and shape estimation datasets, H3.6M \cite{h36m_pami} and 3DPW \cite{von2018recovering:3dpw}. In the task of reconstructing human meshes using 2D detections without depending on paired 3D labels, Key2Mesh outperforms prior models \cite{song2020human:LGD, gong2022cra, choutas2022learning:FMM} in terms of PA-MPJPE performance on both datasets. as well as in MPJPE and PVE on the 3DPW dataset. In addition, thanks to Key2Mesh's simple architecture, which implements a single forward pass, our model is up to $33\times$ faster compared to LGD \cite{song2020human:LGD} -- the only model that (i) solves exactly the same task as Key2Mesh, (ii) is trained on an unpaired dataset, and (iii) has a public implementation.

\section{Related Work}
\label{sec:relatedwork}

\subsection{Human Pose and Shape Estimation}

Considering the existing work on HPS estimation, our model Key2Mesh has four critical properties: (i) For inference, 2D keypoints are used as input, and no RGB data is required. (ii) During training, it does not need a ``paired'' dataset, i.e., 
$\{$image, 3D label$\}$ pairs. It only requires a dataset with 3D body meshes without any corresponding visual data. MoCap datasets provide such data. (iii) During inference, the model runs only a single forward pass, without needing any recurrent or iterative operations. (iv) It uses an adversarial domain adaptation method to adapt the model to visual data. Still, no 3D labels (i.e. paired data) are required. 

Existing methods that have the first two properties above are all iterative: SMPLify \cite{bogo2016keep}, LGD \cite{song2020human:LGD} and NeuralFitter \cite{choutas2022learning:FMM}, while Key2Mesh runs in a single-forward pass. Due to this difference, Key2Mesh is up to 33x faster than LGD, which is reported to be faster than SMPLify. We cannot do any inference time comparison with NeuralFitter as they do not provide code. Another key difference is domain adaptation, which shows its effect in the end result: Key2Mesh outperforms SMPLify, LGD and NeuralFitter in terms of PA-MPJPE on both H3.6M and 3DPW, as well as in MPJPE and PVE on 3DPW.

Several studies utilize unpaired 3D data for training or none, aligning with property (ii). Among these, some require supplementary 2D observations, such as 2D keypoint sequences (PoseNet3D \cite{tripathi2020posenet3d}), dense correspondences (CRA \cite{gong2022cra}), silhouettes (STRAPS \cite{sengupta2020synthetic:STRAP}, Skeleton2Mesh \cite{yu2021skeleton2mesh}), and edge images (STRAPS V3 \cite{sengupta2021hierarchical:STRAPV3}). In contrast, Key2Mesh only requires 2D keypoints as input, making it easy to use and more flexible by eliminating the need for additional 2D estimation. Alternatively, some studies rely on RGB images as inputs such as HMR \cite{kanazawa2018end:HMR} and SPIN \cite{kolotouros2019spin} and report results both using paired and unpaired data. They exhibit a significant performance degradation in unpaired settings, whereas Key2Mesh outperforms them under such conditions. 

A related line of research employs paired data for training, utilizing an RGB image-based pipeline to estimate human pose and shape \cite{kanazawa2018end:HMR, kolotouros2021probabilistic, moon2020i2l, pymaf2021, kocabas2021pare, li2022cliff, ma20233d:VirtualMarkers}. However, obtaining these paired datasets poses challenges. Pose2Mesh \cite{choi2020pose2mesh}, and MPT \cite{lin2024mpt} adopts a hybrid approach, aiming to estimate pose and shape from 2D keypoints similar to (i), combining paired datasets with 2D ground truth and unpaired datasets. Also, MPT utilizes a pre-training strategy with MoCap data similar to ours, but implements an end-to-end fine-tuning pipeline using image-3D ground truth pairs through a 2D pose estimation backbone. In contrast, Key2Mesh leverages only unpaired datasets, implementing domain adaptation to mitigate the discrepancies between 2D ground truths and detections.

\subsection{Unsupervised Domain Adaptation in HPS}
In the realm of HPS estimation, the integration of domain adaptation techniques has recently commenced to address domain gaps. Mugaludi \etal \cite{mugaludi2021aligning} proposed a silhouette based technique to alleviate domain shifts. BOA \cite{guan2021bilevel:BOA}, along with its extension DynaBOA \cite{guan2022out:DynaBOA}, introduced an online domain adaptation method for deploying HPS estimation models in out-of-domain scenarios. They employ a bi-level video-based update rule to adapt the initial model trained on paired 3D datasets. In a different approach, CycleAdapt \cite{mugaludi2021aligning} presented a test-time adaptation framework that does not rely on 2D observations, in contrast to the BOA family. Unlike all these methods, Key2Mesh utilizes 2D keypoints as a proxy representation, relies solely on MoCap data for training the initial model, and implements an efficient single-forward pass estimation pipeline without the need for updating model parameters during inference.

\section{Method}
\label{sec:method}

\begin{figure*}

\centering

\includegraphics[width=0.8\textwidth]{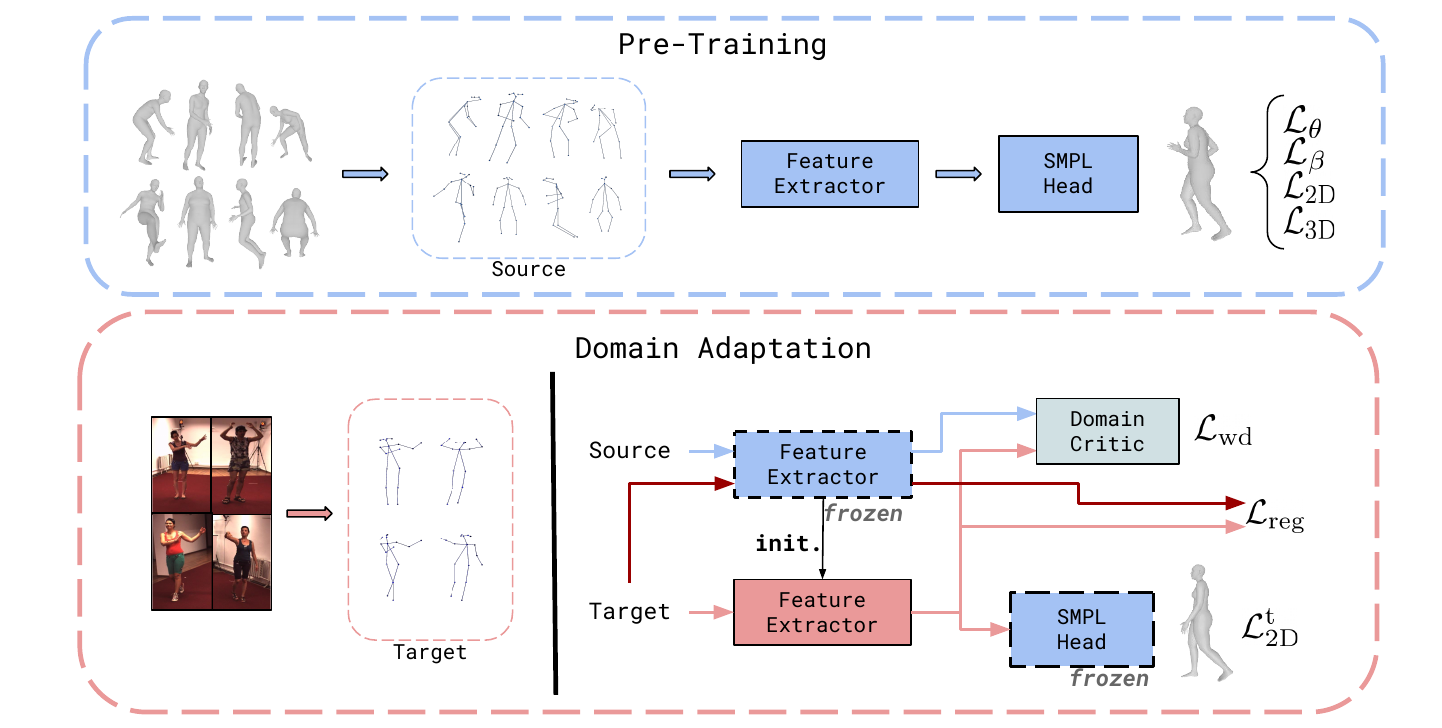}

\caption{\textbf{Overview of our pre-training and domain adaptation phases.} First, we begin with ``Pre-Training’’ phase where we train a feature extractor and an SMPL head using only an unpaired MoCap dataset. Second, we introduce a ``Domain Adaptation’’ phase to close the gap between the source domain (MoCap) and the target domain (Visual) by using 2D detections obtained by an off-the-shelf 2D pose estimator in the target domain.}

\label{fig:main-method}

\end{figure*}

This section describes how we train our Key2Mesh model.
The training is divided into two phases: ``Pre-Training’’ and ``Domain Adaptation’’, as illustrated in \cref{fig:main-method}. In the first phase, we train Key2Mesh using a large MoCap dataset. Then, we introduce a domain adaptation pipeline to adapt our trained model to the testing domain (target).

\subsection{Problem Setting}
\label{ProblemSetting}
Following previous work \cite{song2020human:LGD, choutas2022learning:FMM}, we define our objective as the estimation of the full 3D mesh of human bodies exclusively from 2D keypoint observations, without reliance on any paired datasets. The 3D mesh of a human body is encoded by Skinned Multi-Person Linear(SMPL) \cite{SMPL:2015} model, which is a parametric body model that generates a mesh denoted as ${M\in \mathbb{R}^{6890\times3}}$ based on the provided pose parameters ${\theta \in \mathbb{R}^{24\times3}}$ that include global rotation parameters, and shape parameters $\beta \in \mathbb{R}^{10}$.
The 3D keypoints ${X \in \mathbb{R}^{k\times3}}$, where $k$ is the number keypoints, are defined as a linear combination of the mesh vertices and can be computed as ${X = WM}$ where $W$ is a learned weight matrix ${W \in \mathbb{R}^{k \times 6890}}$. Our goal can be summarized as estimating full 3D human body meshes encoded with SMPL parameters from the 2D keypoint detections obtained by an off-the-shelf 2D pose estimator, \emph{e.g.} OpenPose \cite{cao2017realtime:OpenPose}.

\subsection{Training Data Generation from Unpaired 3D Human Body Data}
\label{sec:data-gen}
 Here we describe the data generation process employed in our first, pre-training, phase. 
 We only utilize publicly available ground truth 3D human body data, obtained by commercial MoCap systems and specialized algorithms like Mosh++ \cite{AMASS:ICCV:2019}, without requiring any corresponding images. 
 We generate pairs of 2D keypoints $x \in \mathbb{R}^{k \times 2}$ and SMPL parameters ($\theta$, $\beta$) on the fly by projecting 3D SMPL keypoints to 2D using a fixed camera $c$ which has a fixed focal length $f$, identity global rotation matrix $R=I^{3\times3}$ looking towards to origin, and fixed translation.
 
Since the 2D keypoints are generated from the ground truth 3D human body, they are free from any defects. However, in the RGB domain, 2D keypoint estimators suffer from problems such as occlusions, noise, image quality, small size of body in pixel space, etc. 
This leads to significant differences between 2D keypoints obtained from MoCap (training)
and those obtained from RGB (test) images. 
To mitigate this disparity, we use a range of augmentation techniques during the training pair generation process. These techniques simulate real-world scenarios observed in RGB
data, incorporating random camera rotations, keypoint occlusions, and the addition of random noise to emulate jitter effects caused by the 2D pose estimators.

\subsection{Pre-Training} 

In the pre-training phase, we train Key2Mesh to regress human pose and shape from 2D keypoints generated from the unpaired dataset, which we refer to as the source dataset. The model consists of a feature extractor and an SMPL head. 

Given 2D keypoints $x$, the feature extractor learns a function $F_\mathrm{pt}: \mathbb{R}^{k\times2} \rightarrow \mathbb{R}^d$ that maps the keypoints to a $d$-dimensional representation $\phi$. Then, SMPL head takes $\phi$ as input and learns a function $H_\mathrm{pt}: \mathbb{R}^d \rightarrow  \mathbb{R}^{\mid \Theta \mid}$ where  $\Theta$ is $[\theta, \beta]$. Following \cite{kolotouros2019spin}, we use a 6D continuous rotation representation \cite{zhou2019continuity:6DRot} in the estimation of SMPL pose parameters $\theta$, instead of axis-angle representation. Finally, for a given $x$, we end up with estimated SMPL parameters $\hat{\theta}$ that includes global rotation parameters, and $\hat{\beta}$.

We compute the estimated 3D keypoints $\hat{X}$ by using the estimated SMPL vertices with learned weight matrix $W$ based on $\hat{\theta}$ and $\hat{\beta}$.  
The estimated 3D keypoints are then projected onto 2D using the fixed camera \(c\) from the training data generation process, yielding the 2D keypoints \(\hat{x}\). 
We further compute losses for SMPL parameters ($\theta$, $\beta$), 2D re-projection ($x$), and 3D ($X$) keypoints and train our overall pre-trained ($F_\mathrm{pt}$, $H_\mathrm{pt}$) model. Specifically, the overall loss function for this phase can be written as follows: 
\begin{equation}\label{eq:DR-Loss}
\mathcal{L}_\mathrm{PT} = w_1\mathcal{L}_{\theta} + w_2\mathcal{L}_{\beta} + w_3\mathcal{L}_\mathrm{2D} + w_4\mathcal{L}_\mathrm{3D}
\end{equation}
where each term is calculated as:
\begin{align*}
\mathcal{L}_{\theta} &= \|\hat{\theta} - \theta \|_1, 
\mathcal{L}_{\beta} = \|\hat{\beta} - \beta \|_1, \\
\mathcal{L}_\mathrm{2D} &= \|\hat{x} - x \|_1,
\mathcal{L}_\mathrm{3D} = \|\hat{X} - X \|_1.
\end{align*}

\subsection{Domain Adaptation}
 
The pre-trained model, trained solely on ground-truth 2D keypoints-SMPL pairs, struggles with domain changes when tested on real indoor and outdoor RGB datasets.
Even though we employed a range of augmentation techniques in the pair generation process to mimic the RGB 2D pose estimator, we observe that the pose variances, cameras, occlusion scenarios, and jitter introduced by the 2D pose estimator are not fully covered by our augmented source domain. Taking inspiration from previous domain adaptation work, ADDA \cite{tzeng2017adversarial:ADDA} and WDGRL \cite{shen2018wasserstein:WDGRL}, we propose an adversarial domain adaptation technique that builds upon our pre-trained model. This technique aims to mitigate the discrepancies between data domains without relying on any labeled data from the target domain. We apply an off-the-shelf pose estimator to the target domain images that are available for use in the training to obtain 2D keypoints which are referred to as the target dataset and use this dataset to adapt our pre-trained model to the target domain.

In this phase, we freeze the parameters of the feature extractor $F_\mathrm{pt}$ and SMPL head $H_\mathrm{pt}$ trained in the previous phase.  Next, we introduce a new feature extractor $F$ which has the same configuration with $F_\mathrm{pt}$ and is initialized with the weights of  $F_\mathrm{pt}$. Additionally, we introduce a domain critic that learns a mapping $D: \mathbb{R}^d \rightarrow \mathbb{R}$ that takes feature vectors from the source and target domain as input and outputs a real number. The domain critic assists us in training $F$ to generate target domain features $\phi^\mathrm{t}$ that are indistinguishable from the features $\phi^\mathrm{s}$ generated by the frozen $F_\mathrm{pt}$ in the source domain, as proposed in WDGRL \cite{shen2018wasserstein:WDGRL}. The final Key2Mesh model closely resembles the pre-trained model, with one key distinction. While the SMPL head remains unchanged, we replace $F_\mathrm{pt}$ with $F$ that has undergone an adaptation process to the target domain. This adaptation process involves utilizing domain adaptation losses, described in detail in  \cref{DomainAdaptationLosses}. By incorporating the adapted feature extractor $F$, we enhance the pipeline's ability to generate accurate SMPL predictions for the given 2D keypoints in the target domain, as the SMPL Head encounters features in the target domain that resemble those from its training set, i.e., the source domain.

\subsection{Domain Critic}
Instead of ADDA's domain discriminator, we follow WDGRL \cite{shen2018wasserstein:WDGRL} and employ a domain critic and calculate Wasserstein distance between the source features $\phi^\mathrm{s}$ generated by $F_\mathrm{pt}$ taken from the pre-trained model and target features $\phi^\mathrm{t}$ generated by $F$ introduced in the adaptation process. The special case of Wasserstein distance between two representation distributions $\mathbb{P}_{\phi^\mathrm{s}}$ and $\mathbb{P}_{\phi^\mathrm{t}}$ proposed in \cite{shen2018wasserstein:WDGRL} can be formulated here as: 
\begin{equation}\label{eq:wasserstein}
    W(\mathbb{P}_{\phi^\mathrm{s}},\mathbb{P}_{\phi^\mathrm{t}}) =
    \sup_{\|D\|_{L} \leq 1}{\mathbb{E}_{\mathbb{P}_{\phi^\mathrm{s}}}[D(\phi)]-\mathbb{E}_{\mathbb{P}_{\phi^\mathrm{t}}}[D(\phi)]},
\end{equation} 
where $\|.\|_{L}$ denotes the Lipschitz constant. When the learned critic function $D$ is enforced to be 1-Lipschitz via  the gradient penalty \cite{gulrajani2017improved:GradPenalty},
we can approximate \cref{eq:wasserstein} with features generated from source samples $x^\mathrm{s}$ and target samples $x^\mathrm{t}$ as:
\begin{equation}\label{L_wd}
    \mathcal{L}_\mathrm{wd}(\Phi^\mathrm{s}, \Phi^\mathrm{t}) = \frac{1}{|x^\mathrm{s}|} \sum\limits_{\phi^\mathrm{s} \in \Phi^\mathrm{s}}  D(\phi^\mathrm{s}) - \frac{1}{|x^\mathrm{t}|}\sum\limits_{\phi^\mathrm{t} \in \Phi^\mathrm{t}} D(\phi^\mathrm{t}),
\end{equation}
where $\Phi^\mathrm{s} = \{ F_\mathrm{pt}(x)\ | \ x \in x^\mathrm{s}\}$ and $\Phi^\mathrm{t} = \{ F(x)\ | \ x \in x^\mathrm{t}\}$. The gradient penalty we use is defined as
\begin{equation}
    \mathcal{L}_\mathrm{grad}(\hat{\phi}) = ( \|\nabla_{\hat{\phi}}D(\hat{\phi}) \|_2 - 1) ^ 2,
\end{equation}
where $\hat{\phi}$ is sampled uniformly along straight lines between source and target feature pairs, similar to \cite{gulrajani2017improved:GradPenalty}.
While training the domain critic, we solve the problem
\begin{equation}
    \max \{ \mathcal{L}_\mathrm{wd} - \gamma \mathcal{L}_\mathrm{grad} \}    
\end{equation}
where $\gamma$ represents the balancing coefficient. Following WDGRL \cite{shen2018wasserstein:WDGRL}, we first train $D$ to optimality with $k$ steps for each domain adaptation training step. Then, by fixing the parameters of optimal $D$, we calculate the domain adaptation losses and update the $F$.

\subsection{Domain Adaptation Losses}\label{DomainAdaptationLosses}
We apply 2D re-projection loss $\mathcal{L}_\mathrm{2D}^\mathrm{t}$, domain loss $\mathcal{L}_\mathrm{wd}$ and regularization on domain-adapted features $\mathcal{L}_\mathrm{reg}$ in the adaptation process. Below, we describe these loss and regularization terms.

\noindent \textbf{2D re-projection loss on the target domain.}
$F$ takes 2D keypoint detections $x^\mathrm{t}$ from the target dataset as inputs, which allows us to calculate a 2D re-projection loss between the final SMPL predictions and $x^\mathrm{t}$. This 2D re-projection loss serves as a supervision signal during the adaptation without requiring human-labeled or 3D ground truth data. Similar to $\mathcal{L}_\mathrm{2D}$, we use the fixed camera $c$ to project predicted 3D SMPL keypoints and obtain 2D keypoints ($\hat{x}^\mathrm{t}$). The calculation of $\mathcal{L}_\mathrm{2D}^\mathrm{t}$ is as follows:
\begin{equation}
    \mathcal{L}_\mathrm{2D}^\mathrm{t} = \|\hat{x}^\mathrm{t} - x^\mathrm{t} \|_1.
\end{equation}

\noindent \textbf{Domain loss.} By utilizing $D$, we enforce $\phi^\mathrm{t}$ to be inseparable from $\phi^\mathrm{s}$. To do this, we solve the problem $ \min\{ \mathcal{L}_\mathrm{wd} \}$. Practically, we re-calculate the $\mathcal{L}_\mathrm{wd}$ after fixing the optimal parameters of the domain critic.

\noindent \textbf{Regularization on domain-adapted features.}
We apply a regularization term on $\phi^\mathrm{t}$. We compute the corresponding feature vector $\bar\phi^\mathrm{t}= F_\mathrm{pt}(x^{\mathrm{t}})$ and apply the following loss:
\begin{equation}
    \mathcal{L}_\mathrm{reg} = \| \phi^\mathrm{t} - \bar\phi^\mathrm{t} \|.
\end{equation}

We combine the loss functions mentioned in this section using specific weights to obtain an overall loss function to train $F$:
\begin{equation}
    \mathcal{L}_\mathrm{DA} = w_5\mathcal{L}_\mathrm{2D}^\mathrm{t} + w_6\mathcal{L}_\mathrm{wd} + w_7\mathcal{L}_\mathrm{reg}. 
\end{equation}

\section{Experiments}
\label{sec:experiments}

\subsection{Implementation Details}
We employ standard MLP architectures: 5 blocks for feature extractors, 2 for the SMPL head. Each block includes a linear layer (1024 hidden dimensions), batch normalization, parametric ReLU, and 0.2 dropout. The Domain Critic uses 2 blocks without normalization layers. For pre-training, we set hyperparameters ($w_1 = w_2 = 100$, $w_3 = w_4 = 50$) for 
$\mathcal{L}_\mathrm{PT}$, 
optimizing it via Adam optimizer \cite{kingma2014adam} (lr: 1e-3) over ten epochs, batch size 256. During domain adaptation, both the domain critic and feature extractor are trained for five epochs, using lr 1e-4, batch size 256 per domain (512 for critic). The domain critic trains $k=3$ times per adaptation step. For $\mathcal{L}_{\mathrm{DA}}$, we set hyperparameters ($w_5=10$, $w_6=0.1$, $w_7=20$). Experiments are conducted using PyTorch \cite{paszke2019pytorch} on one RTX3080ti GPU for all experiments.

\subsection{Datasets and Evaluation Metrics}

\subsubsection{Training Data}
During the pre-training of Key2Mesh (Fig. \ref{fig:main-method}), we solely utilize the AMASS dataset \cite{AMASS:ICCV:2019}. This extensive 3D MoCap dataset consists of a wide range of human poses and shapes (and motions), with SMPL parameters. However, it does not include RGB images corresponding to the 3D data.

In the domain adaptation phase, we leverage the InstaVariety dataset \cite{humanMotionKZFM19:instavariety}, which offers a large set of in-the-wild 2D keypoint detections. Moreover, our domain adaptation process incorporates the Human3.6M dataset \cite{h36m_pami} (H3.6M) and the 3DPW dataset \cite{von2018recovering:3dpw}.
\textbf{We do not use any ground-truth labels available in these datasets.} Instead, we employ OpenPose \cite{cao2017realtime:OpenPose} to obtain detected 2D keypoints for H3.6M \cite{h36m_pami} and the detections available in 3DPW \cite{von2018recovering:3dpw}. We use these 2D detections in the adaptation phase. 

\subsubsection{Evaluation Data}
We quantitatively and qualitatively evaluate our approach on widely used indoor and in-the-wild datasets, H3.6M \cite{h36m_pami} (subjects S9, S11) and 3DPW \cite{von2018recovering:3dpw} datasets.

We use OpenPose \cite{cao2017realtime:OpenPose} on the H3.6M test subjects to obtain 2D keypoint detections, and following the evaluation protocol P2, we only used a frontal camera (camera 3). Following the approach of prior works \cite{song2020human:LGD, choutas2022learning:FMM}, we discard frames in which fewer than 6 keypoints were detected. For the 3DPW evaluation, we use the detections available in the dataset. We report PA-MPJPE and MPJPE for the H3.6M dataset and PA-MPJPE, MPJPE and PVE for the 3DPW dataset, all measured in mm.

\subsection{Quantitative Results}

\begin{table}[]
\centering
\begin{adjustbox}{max width=\columnwidth}
\begin{tabular}{|c|c|c|c|c|}
\hline
Method & Input & PA-MPJPE$\downarrow$ & MPJPE$\downarrow$ & Time$\downarrow$ \\ \hline \hline
SMPLify \cite{bogo2016keep} & $K$ & 82.3 & - & - \\ \hline
HMR (\textit{u}) \cite{kanazawa2018end:HMR} & $I$ & 66.5 & - & - \\  \hline
SPIN (\textit{u}) \cite{kolotouros2019spin} & $I$ & 62.0 & - & - \\ \hline
STRAPS \cite{sengupta2020synthetic:STRAP} & $K, S$ & 55.4 & - & -  \\ \hline
LGD \cite{song2020human:LGD} & $K$ & 55.0 & 102.4 & 42.4 \\ \hline
CRA \cite{gong2022cra} & $K, D$ & 53.9 & \textbf{81.0} & ($D$: 110.0)+ \\ \hline \hline
Key2Mesh\textsuperscript{\dag} & $K$ & 51.4 & 108.1 &\textbf{3.4} \\ \hline
Key2Mesh\textsuperscript{\ddag} & $K$ & \textbf{51.0} & 107.1 & \textbf{3.4} \\ \hline
\end{tabular}
\end{adjustbox}
\caption{\textbf{Comparison of models trained on \underline{unpaired data} similar to ours, assessed on H3.6M \cite{h36m_pami} for PA-MPJPE and MPJPE (both in mm) under P2, and processing time (ms) with a single batch size.}
Inputs include ($D$)ense Correspondences, ($I$)mage, 2D ($K$)eypoints, or ($S$)ilhouette. 
As CRA does not have a publicly available code,
our time evaluation focused on measuring the time needed to acquire  dense correspondences ($D$) (DensePose \cite{guler2018densepose}), an input used alongside keypoints ($K$) unlike Key2Mesh and LGD. (\textit{u}) denotes unpaired. Our models marked with (\dag) and (\ddag) were obtained by adapting our PT model to training subjects and evaluation subjects of the H3.6M dataset, respectively, without employing any ground truth labels. 
Our Key2Mesh obtains the best PA-MPJPE result and is $12\times$ faster than LGD and at least $32\times$ faster than CRA.}
\label{table:h36m}
\end{table}

\begin{table}[]
\begin{adjustbox}{max width=\columnwidth}
\begin{tabular}{|c|c|c|c|c|}
\hline
Method & Input & PA-MPJPE$\downarrow$ & MPJPE$\downarrow$ & PVE$\downarrow$\\ \hline \hline
SMPLify \cite{bogo2016keep} & $K$ & 106.1 & - & -\\ \hline
PoseNet3D\textsuperscript{*} \cite{tripathi2020posenet3d} & $K$ & 73.6 & - & -\\ \hline
STRAPS V3 \cite{sengupta2021hierarchical:STRAPV3} & $K, E$ & 59.2 & - & -\\ \hline
CRA \cite{gong2022cra} & $K, D$ & 55.9 & 89.1 & 115.3 \\ \hline
LGD \cite{song2020human:LGD} & $K$ & 54.2 & 91.3 & 104.8 \\ \hline
NeuralFitter \cite{choutas2022learning:FMM} & $K$ & 52.2 & - & - \\ \hline \hline 
Key2Mesh\textsuperscript{\dag} & $K$ & 50.1 & 89.0 & 101.7 \\ \hline
Key2Mesh\textsuperscript{\ddag} & $K$ & \textbf{49.8} & \textbf{86.7} & \textbf{99.5} \\ \hline
\end{tabular}
\end{adjustbox}
\centering
\caption{\textbf{Comparison of our models with others that utilize \underline{2D proxy representations} and trained on \underline{unpaired data} similar to our model like ours, assessed on the 3DPW \cite{von2018recovering:3dpw} dataset for PA-MPJPE, MPJPE and PVE (all in mm).} The input for the presented methods includes ($D$)ense Correspondences, ($E$)dge Images, and 2D ($K$)eypoints. (*) denotes methods that take temporal data as  input.
(\dag) and (\ddag) were achieved through domain adaptation of the PT model to the InstaVariety dataset and the test split of the 3DPW dataset, respectively, without utilizing ground-truth labels.}
\label{table:3dpw}
\end{table}

We first pre-train Key2Mesh by following the data generation process detailed in \cref{sec:data-gen}. Then, we apply our domain adaptation method by using OpenPose detections from H3.6M subjects, 3DPW, and InstaVariety. For H3.6M, we report the performance of the domain-adapted model, in which we used the detections from all H3.6M training subjects to adapt our pre-trained model. For the 3DPW dataset, we report our InstaVariety adapted model since adaptation to InstaVariety leads to better results compared to the 3DPW train split. Our adaptation process doesn't require target labels, allowing direct adaptation of the pre-trained model to the test sets using the respective OpenPose detections. Consequently, we present evaluations for models adapted to both H3.6M evaluation subjects and the 3DPW test split. We present detailed evaluations on the target dataset in \cref{sec:target-dataset-size}.

\noindent \textbf{Evaluation on H3.6M.} We compare our models with others on H3.6M benchmark under Protocol 2 (P2). Since we only use 2D keypoint observations without relying on images and utilize unpaired 3D training data, it is only possible to compare Key2Mesh with other models following a similar approach. As presented in \cref{table:h36m}, our model outperforms existing state-of-the-art methods in terms of PA-MPJPE without relying on additional 2D observations like dense correspondences or silhouettes used in CRA \cite{gong2022cra} and STRAPS \cite{sengupta2020synthetic:STRAP}. However, our method falls short in MPJPE compared to CRA, largely because CRA leverages dense correspondences for estimation. This advantage is more pronounced in controlled environments like H3.6M, where obtaining high-quality dense correspondences is less challenging. For a fair comparison, we evaluate an iterative baseline LGD \cite{song2020human:LGD} on our 2D detections, as the original study did not utilize OpenPose \cite{cao2017realtime:OpenPose} keypoint detections (\cref{table:h36m} row 5). Note that we achieve improved PA-MPJPE for LGD using our 2D detections compared to what its authors report.
While our model improves upon iterative methods such as LGD and SMPLify \cite{bogo2016keep} in PA-MPJPE, it demonstrates relatively lower performance in MPJPE when compared to LGD. 
Our model is $12\times$ faster than LGD and at least $32\times$ faster than CRA.
In addition, we compared our model with some image-based baselines, HMR \cite{kanazawa2018end:HMR} and SPIN \cite{kolotouros2019spin}, under the unpaired training setting. our model outperforms these baselines.

\noindent \textbf{Evaluation on 3DPW.} In \cref{table:3dpw}, we report PA-MPJPE, MPJPE and PVE on 3DPW test set. We compare our model with others that utilize 2D proxy representations such as dense correspondences, 2D keypoints, and edge images to recover SMPL meshes. Also, similar to ours, these models only rely on unpaired data during training. Noticeably, prior works such as CRA \cite{gong2022cra}, STRAPSV3 \cite{sengupta2021hierarchical:STRAPV3} and Posenet3D \cite{tripathi2020posenet3d}, as depicted in the \cref{table:3dpw}, require additional inputs or temporal keypoint detections to regress SMPL meshes.
Our model achieves better performance without such dependencies for in-the-wild scenarios.
We also compare our model with iterative ones such as SMPLify \cite{bogo2016keep}, LGD \cite{song2020human:LGD}, and NeuralFitter \cite{choutas2022learning:FMM}, by evaluating  LGD in MPJPE and PVE (metrics initially unreported), we test LGD on the 3DPW dataset. This evaluation yields a slight improvement in LGD's PA-MPJPE compared to its original reported values, and we obtain MPJPE and PVE metrics for comparison. Across all metrics, our model outperforms these iterative methods.

\noindent \textbf{Processing time comparison.} We also compared our model's processing time with LGD and CRA in \cref{table:h36m}. As CRA lacks available code, our time evaluation focused on measuring the time needed to acquire dense correspondences (DensePose \cite{guler2018densepose} with ResNet-101 backbone), an extra input used alongside keypoints, unlike Key2Mesh and LGD. Our inference pipeline is highly efficient, operating at 3.4 ms with a batch size of 1, compared to LGD's 42.4 ms where DensePose requires 110.0 ms for Dense Correspondences on the same hardware. With a larger batch size of 512, our method achieves 3.6 ms, a speed advantage of 33$\times$ faster than LGD, which slows down to 120 ms due to its iterative nature. These comparisons highlight our method's crucial rapid processing for speed-focused applications.

\subsection{Ablation Studies}
\label{sec:ablation}

\subsubsection{Ablation on Domain Adaptation Loss Terms}
\begin{table}
\centering
\begin{adjustbox}{max width=0.79\columnwidth}
\begin{tabular}{|cccc|}
\hline
\multicolumn{2}{|c}{Losses} & PA-MPJPE$\downarrow$ & MPJPE$\downarrow$ \\ \hline \hline
\multicolumn{2}{|c}{Pre-Trained} & 53.9 & 114.7 \\ \hline
\multirow{3}{*}{DA} & $\mathcal{L}_\mathrm{2D}^\mathrm{t}$ & 53.8 & 109.0 \\ \cline{2-4} 
 & $\mathcal{L}_\mathrm{2D}^\mathrm{t}$ + $\mathcal{L}_\mathrm{wd}$ & 52.2 & 109.0 \\ \cline{2-4} 
 & $\mathcal{L}_\mathrm{2D}^\mathrm{t}$ + $\mathcal{L}_\mathrm{wd}$ + $\mathcal{L}_\mathrm{reg}$ & \textbf{51.4} & \textbf{108.1} \\ \cline{1-4} 
\end{tabular}
\end{adjustbox}
\caption{\textbf{Ablation on domain adaptation losses.} We conduct experiments to assess the impact of loss terms during the domain adaptation phase using the H3.6M \cite{h36m_pami} dataset. We start with a pre-trained model and add (D)omain (A)daptation losses.}
\label{table:loss-ablation}
\end{table}

We conduct experiments to measure the effects of the loss terms during the domain adaptation phase. As shown in \cref{table:loss-ablation}, starting from a pre-trained model on the AMASS dataset, we employ our domain adaptation method using varying combinations of loss terms such as $\mathcal{L}_\mathrm{2D}^\mathrm{t}$, $\mathcal{L}_\mathrm{wd}$, $\mathcal{L}_\mathrm{reg}$. Applying domain adaptation with only 2D re-projection loss on the target ($\mathcal{L}_\mathrm{2D}^\mathrm{t}$) helps the model to better align its predictions on the target domain. This approach leads to a slight enhancement in the PA-MPJPE metric and a noticeable improvement in MPJPE. We obtain further improvement in PA-MPJPE when we use a combination of $\mathcal{L}_\mathrm{2D}^\mathrm{t}$ and domain critic loss ($\mathcal{L}_\mathrm{wd}$). This observation supports our intuition that adapting the features computed on the target dataset helps in reducing pose estimation errors. Also, by introducing a regularization term on the domain-adapted features ($\mathcal{L}_\mathrm{reg}$), we achieved the best PA-MPJPE and MPJPE. This indicates that applying regularization to the domain-adapted features leads to a more effective feature extraction for the target dataset.

\subsubsection{Ablation on the Target Dataset and Size}
\label{sec:target-dataset-size}
\begin{table}
\centering
\begin{adjustbox}{max width=\columnwidth}
\begin{tabular}{|c|c|c|c|c|}
\hline
Method & Target Dataset & Size & PA-MPJPE$\downarrow$ & MPJPE$\downarrow$\\ \hline \hline
Ours - PT & - & - & 53.9 & 114.7 \\ \hline
Ours - DA & InstaVariety & 2.19M & 52.0 & 111.3 \\ \hline
Ours - DA & H3.6M - S1 & 23.5K & 52.8 & 108.9 \\ \hline
Ours - DA & H3.6M - Train\textsuperscript{*} & 160K & 51.4 & 108.1 \\ \hline
Ours - DA & H3.6M - Val.\textsuperscript{\dag} & 12.4K & \textbf{51.0} & \textbf{107.1} \\ \hline
\end{tabular}
\end{adjustbox}
\caption{\textbf{Ablation on target dataset and size using H3.6M \cite{h36m_pami}.} PT and DA denote the pre-trained model and the domain-adapted model, respectively. (*) represents training subjects (S1, S5-S8) and (\dag) indicates validation subjects (S9, S11).}
\label{table:h36m-target-ablation}
\end{table}

\begin{table}[]
\centering
\begin{adjustbox}{max width=\columnwidth}
\begin{tabular}{|c|c|c|c|c|c|}
\hline
Method & Target Dataset & Size & PA$\downarrow$ & MPJPE$\downarrow$ & PVE$\downarrow$ \\ \hline \hline
Ours - PT & - & - & 57.5 & 93.6 & 109.8 \\ \hline
Ours - DA & 3DPW-Train & 22.8K & 57.4 & 91.7 & 107.3 \\ \hline
Ours - DA & InstaVariety & 2.19M & 57.1 & 91.7 & 107.1 \\ \hline
Ours - DA & 3DPW-Val. & 33.5K & \textbf{56.4} & \textbf{88.6} & \textbf{104.4} \\ \hline
\end{tabular}
\end{adjustbox}
\caption{\textbf{Ablation on target dataset and size using 3DPW \cite{von2018recovering:3dpw} validation set.} PT and DA represent the pre-trained model and the domain-adapted model, respectively, with PA referring to PA-MPJPE.}
\label{table:3dpw-target-ablation}
\end{table}

We conduct measurements to assess the impact of the target dataset and size during our domain adaptation phase, using H3.6M and 3DPW validation sets. By adapting our pre-trained model to different target datasets, including validation sets, we calculate the evaluation metrics for each adapted model.

\noindent \textbf{Results on H3.6M.}  For H3.6M, we adapt our pre-trained model to four different datasets separately: InstaVariety, a small dataset that includes only a single subject (S1) from the H3.6M training subjects, the entire H3.6M training subjects, and validation subjects. \cref{table:h36m-target-ablation} row 2 shows that despite InstaVariety being an in-the-wild 2D dataset, different from H3.6M, our domain adaptation process reduces pose alignment errors. In row 3, even using a single subject shows a reduction in error metrics thanks to our domain adaptation process. Row 4 shows the model adapted using the entire H3.6M training subjects, demonstrating better performance than the model adapted using only a single subject.
Lastly, by employing our label-independent domain adaptation framework, we adapt our pre-trained model to the validation subjects, resulting in the best PA-MPJPE and MPJPE compared to all other adaptation strategies. Interestingly, the model adapted using the H3.6M training subjects performs comparably to the one adapted specifically to validation subjects. This underlines the potential of using pre-collected data for controlled datasets like H3.6M.

\noindent \textbf{Results on 3DPW.} We leverage InstaVariety, 3DPW-Train, and 3DPW-Validation datasets in the adaptation process for the 3DPW dataset. In \cref{table:3dpw-target-ablation}, row 2 demonstrates that incorporating the 3DPW-train dataset, which is relatively small, leads to improvements in the evaluation metrics. In addition, \cref{table:3dpw-target-ablation} row 3 shows that using the InstaVariety dataset, which is also an in-the-wild dataset,  further reduces PA-MPJPE. Finally, we employed the 3DPW-validation dataset to enhance our model, leveraging our label-independent adaptation framework, resulting in the best metrics (\cref{table:3dpw-target-ablation} row 4). This implies that in a challenging in-the-wild dataset like 3DPW, direct adaptation to the test data results in more accurate pose alignment.

\begin{figure}
\centering
\includegraphics[width=0.47\textwidth]{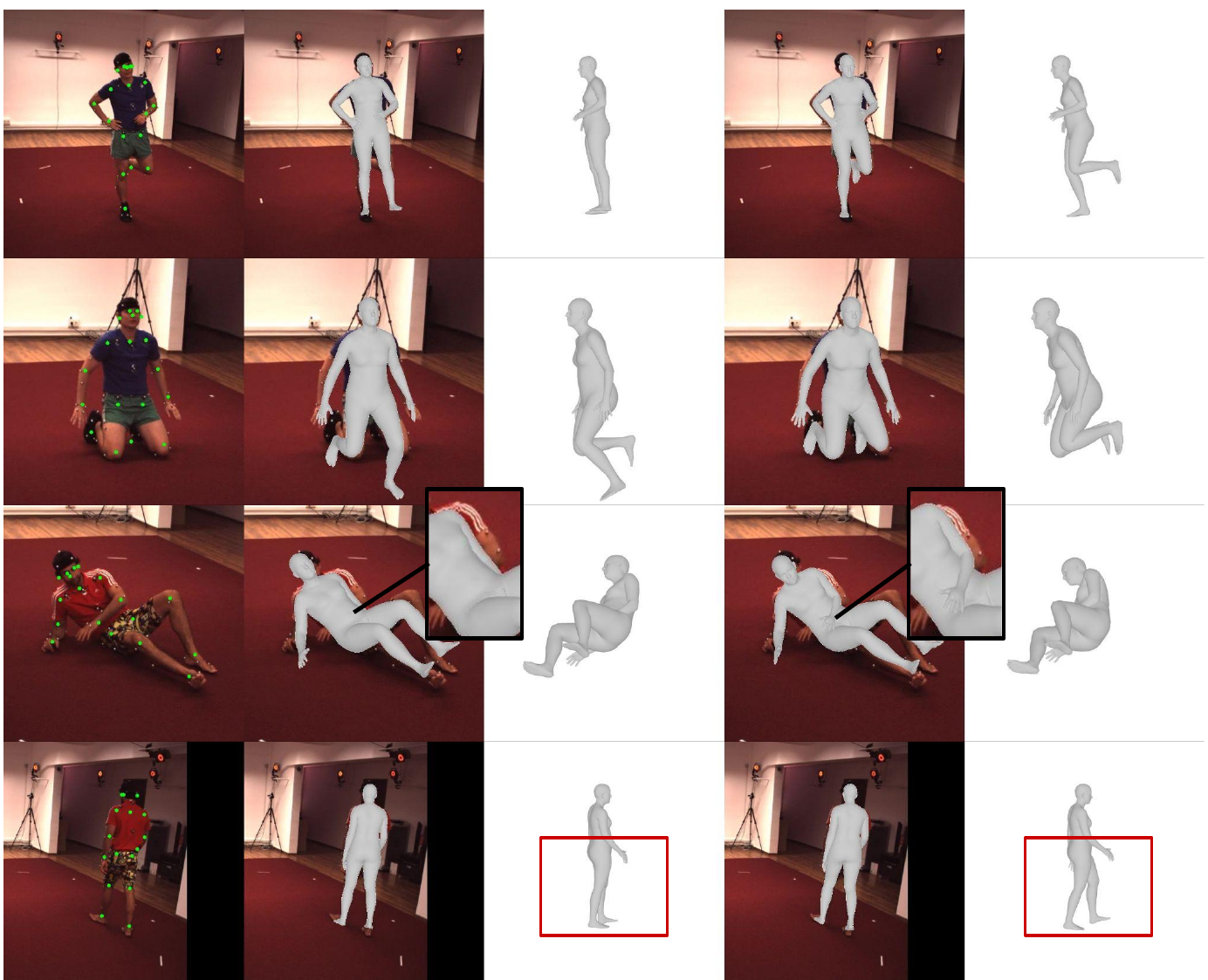}
\caption{\textbf{Qualitative comparison of our pre-trained and domain-adapted model on H3.6M dataset \cite{h36m_pami}.} The first column shows the input image and keypoint detections. The second and third columns display outputs from the pre-trained model, while the last two columns present results from the domain-adapted model, which outperforms our pre-trained model, especially on complex poses.}
\label{fig:h36m-ablation}
\end{figure}

\subsection{Qualitative Results}

\begin{figure}
\centering
\includegraphics[width=0.47\textwidth]{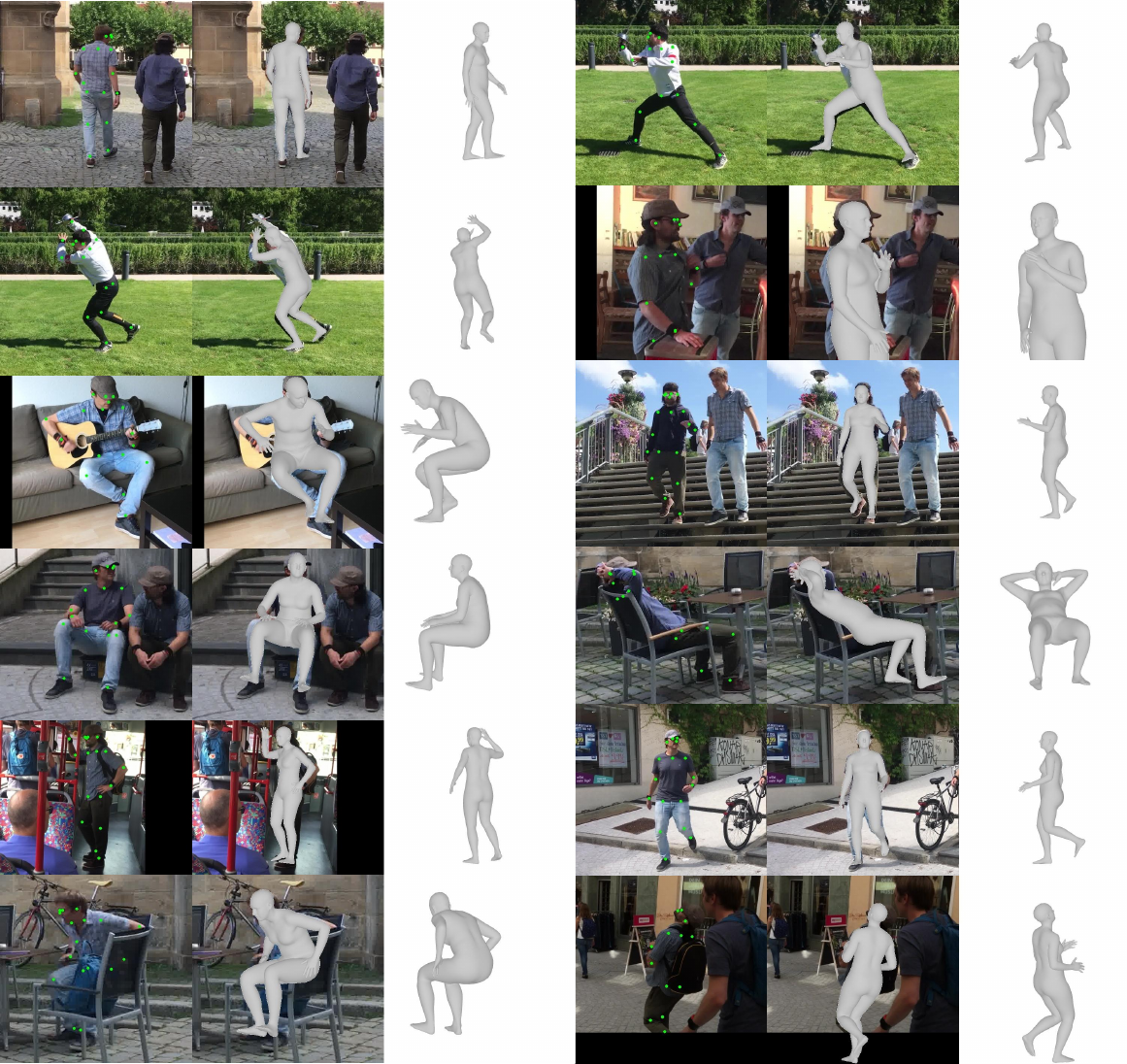}
\caption{\textbf{Qualitative results on the 3DPW dataset \cite{von2018recovering:3dpw}.} Each sample displays the input image and its corresponding 2D keypoint detections in the first column, the SMPL mesh overlaid on the image in the second column, and the side view of the SMPL mesh in the third column, all generated using a domain-adapted model.}
\label{fig:3dpw-qual}
\end{figure}

Our method does not involve estimating camera translation.
To produce qualitative results, 
we compute the best camera translation using a least squares optimization to overlay meshes onto images. This optimization minimizes the difference between the 3D keypoints of the SMPL meshes, projected onto the image plane, and their corresponding 2D keypoint inputs.

In \cref{fig:h36m-ablation}, we qualitatively compare our pre-trained and adapted models. The results show that domain adaptation helps our model to produce more aligned results. Further qualitative results showcasing our domain-adapted models for the 3DPW dataset are presented in \cref{fig:3dpw-qual}.

\section{Conclusion}
\label{sec:conclusion}
In this study, we present Key2Mesh for recovering 3D human mesh parameters from 2D keypoints. We address the lack of image-to-3D, ``paired’’ training data and utilize an ``unpaired’’ training setting. We introduce an adversarial domain adaptation method to adapt models trained on unpaired data to visual data. We comprehensively evaluate our Key2Mesh model on widely used indoor and in-the-wild benchmarks, H3.6M and 3DPW datasets. Our model outperforms similar methods,  showcasing superior PA-MPJPE performance on both datasets, along with superior MPJPE and PVE performance on the 3DPW dataset. Comparatively, our model is significantly faster (up to 33$\times$) than the previous state-of-the-art model LGD \cite{song2020human:LGD}, emphasizing its efficiency for applications where processing time is critical.
\\ \\
\small{\noindent \textbf{Acknowledgements} Dr. Akbas is supported by the ``Young Scientist Awards Program (BAGEP)'' of Science Academy, Turkey.}
{
    \small
    \bibliographystyle{ieeenat_fullname}
    \bibliography{main}

\begin{thebibliography}{57}
\providecommand{\natexlab}[1]{#1}
\providecommand{\url}[1]{\texttt{#1}}
\expandafter\ifx\csname urlstyle\endcsname\relax
  \providecommand{\doi}[1]{doi: #1}\else
  \providecommand{\doi}{doi: \begingroup \urlstyle{rm}\Url}\fi

\bibitem[Bogo et~al.(2016)Bogo, Kanazawa, Lassner, Gehler, Romero, and Black]{bogo2016keep}
Federica Bogo, Angjoo Kanazawa, Christoph Lassner, Peter Gehler, Javier Romero, and Michael~J Black.
\newblock Keep it smpl: Automatic estimation of 3d human pose and shape from a single image.
\newblock In \emph{European Conference on Computer Vision}, pages 561--578. Springer, 2016.

\bibitem[Cao et~al.(2017)Cao, Simon, Wei, and Sheikh]{cao2017realtime:OpenPose}
Zhe Cao, Tomas Simon, Shih-En Wei, and Yaser Sheikh.
\newblock Realtime multi-person 2d pose estimation using part affinity fields.
\newblock In \emph{Proceedings of the IEEE Conference on Computer Vision and Pattern Recognition}, pages 7291--7299, 2017.

\bibitem[Chen et~al.(2018)Chen, Gabriel, Alasfour, Gong, Doyle, Devinsky, Friedman, Dugan, Melloni, Thesen, et~al.]{chen2018patient}
Kenny Chen, Paolo Gabriel, Abdulwahab Alasfour, Chenghao Gong, Werner~K Doyle, Orrin Devinsky, Daniel Friedman, Patricia Dugan, Lucia Melloni, Thomas Thesen, et~al.
\newblock Patient-specific pose estimation in clinical environments.
\newblock \emph{IEEE journal of translational engineering in health and medicine}, 6:\penalty0 1--11, 2018.

\bibitem[Cho et~al.(2023)Cho, Cho, Ahn, and Kim]{cho2023implicit}
Hanbyel Cho, Yooshin Cho, Jaesung Ahn, and Junmo Kim.
\newblock Implicit 3d human mesh recovery using consistency with pose and shape from unseen-view.
\newblock In \emph{Proceedings of the IEEE/CVF Conference on Computer Vision and Pattern Recognition}, pages 21148--21158, 2023.

\bibitem[Cho et~al.(2022)Cho, Youwang, and Oh]{cho2022FastMETRO}
Junhyeong Cho, Kim Youwang, and Tae-Hyun Oh.
\newblock Cross-attention of disentangled modalities for 3d human mesh recovery with transformers.
\newblock In \emph{European Conference on Computer Vision (ECCV)}, 2022.

\bibitem[Choi et~al.(2020)Choi, Moon, and Lee]{choi2020pose2mesh}
Hongsuk Choi, Gyeongsik Moon, and Kyoung~Mu Lee.
\newblock Pose2mesh: Graph convolutional network for 3d human pose and mesh recovery from a 2d human pose.
\newblock In \emph{European Conference on Computer Vision}, pages 769--787. Springer, 2020.

\bibitem[Choi et~al.(2021)Choi, Moon, Chang, and Lee]{choi2020beyond}
Hongsuk Choi, Gyeongsik Moon, Ju~Yong Chang, and Kyoung~Mu Lee.
\newblock Beyond static features for temporally consistent 3d human pose and shape from a video.
\newblock In \emph{Conference on Computer Vision and Pattern Recognition}, 2021.

\bibitem[Choutas et~al.(2022)Choutas, Bogo, Shen, and Valentin]{choutas2022learning:FMM}
Vasileios Choutas, Federica Bogo, Jingjing Shen, and Julien Valentin.
\newblock Learning to fit morphable models.
\newblock In \emph{European Conference on Computer Vision}, pages 160--179. Springer, 2022.

\bibitem[Gong et~al.(2022)Gong, Zheng, Planche, Karanam, Chen, Doermann, and Wu]{gong2022cra}
Xuan Gong, Meng Zheng, Benjamin Planche, Srikrishna Karanam, Terrence Chen, David Doermann, and Ziyan Wu.
\newblock Self-supervised human mesh recovery with cross-representation alignment.
\newblock In \emph{European Conference on Computer Vision}, pages 212--230, 2022.

\bibitem[Grimm et~al.(2012)Grimm, Bauer, Sukkau, Hornegger, and Greiner]{grimm2012markerless}
Robert Grimm, Sebastian Bauer, Johann Sukkau, Joachim Hornegger, and G{\"u}nther Greiner.
\newblock Markerless estimation of patient orientation, posture and pose using range and pressure imaging: For automatic patient setup and scanner initialization in tomographic imaging.
\newblock \emph{International journal of computer assisted radiology and surgery}, 7:\penalty0 921--929, 2012.

\bibitem[Guan et~al.(2021)Guan, Xu, Wang, Ni, and Yang]{guan2021bilevel:BOA}
Shanyan Guan, Jingwei Xu, Yunbo Wang, Bingbing Ni, and Xiaokang Yang.
\newblock Bilevel online adaptation for out-of-domain human mesh reconstruction.
\newblock In \emph{Proceedings of the IEEE Conference on Computer Vision and Pattern Recognition}, pages 10472--10481, 2021.

\bibitem[Guan et~al.(2022)Guan, Xu, He, Wang, Ni, and Yang]{guan2022out:DynaBOA}
Shanyan Guan, Jingwei Xu, Michelle~Zhang He, Yunbo Wang, Bingbing Ni, and Xiaokang Yang.
\newblock Out-of-domain human mesh reconstruction via dynamic bilevel online adaptation.
\newblock \emph{IEEE Transactions on Pattern Analysis and Machine Intelligence}, 45\penalty0 (4):\penalty0 5070--5086, 2022.

\bibitem[G{\"u}ler et~al.(2018)G{\"u}ler, Neverova, and Kokkinos]{guler2018densepose}
R{\i}za~Alp G{\"u}ler, Natalia Neverova, and Iasonas Kokkinos.
\newblock Densepose: Dense human pose estimation in the wild.
\newblock In \emph{Proceedings of the IEEE Conference on Computer Vision and Pattern Recognition}, pages 7297--7306, 2018.

\bibitem[Gulrajani et~al.(2017)Gulrajani, Ahmed, Arjovsky, Dumoulin, and Courville]{gulrajani2017improved:GradPenalty}
Ishaan Gulrajani, Faruk Ahmed, Martin Arjovsky, Vincent Dumoulin, and Aaron~C Courville.
\newblock Improved training of wasserstein gans.
\newblock \emph{Advances in Neural Information Processing Systems}, 30, 2017.

\bibitem[Guzov et~al.(2021)Guzov, Mir, Sattler, and Pons-Moll]{guzov2021human}
Vladimir Guzov, Aymen Mir, Torsten Sattler, and Gerard Pons-Moll.
\newblock Human poseitioning system (hps): 3d human pose estimation and self-localization in large scenes from body-mounted sensors.
\newblock In \emph{Proceedings of the IEEE Conference on Computer Vision and Pattern Recognition}, pages 4318--4329, 2021.

\bibitem[Ionescu et~al.(2014)Ionescu, Papava, Olaru, and Sminchisescu]{h36m_pami}
Catalin Ionescu, Dragos Papava, Vlad Olaru, and Cristian Sminchisescu.
\newblock Human3.6m: Large scale datasets and predictive methods for 3d human sensing in natural environments.
\newblock \emph{IEEE Transactions on Pattern Analysis and Machine Intelligence}, 36\penalty0 (7):\penalty0 1325--1339, 2014.

\bibitem[Joo et~al.(2020)Joo, Neverova, and Vedaldi]{joo2020eft}
Hanbyul Joo, Natalia Neverova, and Andrea Vedaldi.
\newblock Exemplar fine-tuning for 3d human pose fitting towards in-the-wild 3d human pose estimation.
\newblock In \emph{International Conference on 3D Vision}, 2020.

\bibitem[Kanazawa et~al.(2018)Kanazawa, Black, Jacobs, and Malik]{kanazawa2018end:HMR}
Angjoo Kanazawa, Michael~J Black, David~W Jacobs, and Jitendra Malik.
\newblock End-to-end recovery of human shape and pose.
\newblock In \emph{Proceedings of the IEEE Conference on Computer Vision and Pattern Recognition}, pages 7122--7131, 2018.

\bibitem[Kanazawa et~al.(2019)Kanazawa, Zhang, Felsen, and Malik]{humanMotionKZFM19:instavariety}
Angjoo Kanazawa, Jason~Y. Zhang, Panna Felsen, and Jitendra Malik.
\newblock Learning 3d human dynamics from video.
\newblock In \emph{Proceedings of the IEEE Conference on Computer Vision and Pattern Recognition}, 2019.

\bibitem[Kingma and Ba(2014)]{kingma2014adam}
Diederik~P Kingma and Jimmy Ba.
\newblock Adam: A method for stochastic optimization.
\newblock \emph{arXiv preprint arXiv:1412.6980}, 2014.

\bibitem[Kocabas et~al.(2019)Kocabas, Karagoz, and Akbas]{kocabas2019epipolar}
Muhammed Kocabas, Salih Karagoz, and Emre Akbas.
\newblock Self-supervised learning of 3d human pose using multi-view geometry.
\newblock In \emph{Proceedings of the IEEE Conference on Computer Vision and Pattern Recognition}, 2019.

\bibitem[Kocabas et~al.(2020)Kocabas, Athanasiou, and Black]{kocabas2019vibe}
Muhammed Kocabas, Nikos Athanasiou, and Michael~J. Black.
\newblock Vibe: Video inference for human body pose and shape estimation.
\newblock In \emph{Proceedings of the IEEE Conference on Computer Vision and Pattern Recognition}, 2020.

\bibitem[Kocabas et~al.(2021)Kocabas, Huang, Hilliges, and Black]{kocabas2021pare}
Muhammed Kocabas, Chun-Hao~P Huang, Otmar Hilliges, and Michael~J Black.
\newblock Pare: Part attention regressor for 3d human body estimation.
\newblock In \emph{Proceedings of the IEEE International Conference on Computer Vision}, pages 11127--11137, 2021.

\bibitem[Kolotouros et~al.(2019)Kolotouros, Pavlakos, Black, and Daniilidis]{kolotouros2019spin}
Nikos Kolotouros, Georgios Pavlakos, Michael~J Black, and Kostas Daniilidis.
\newblock Learning to reconstruct 3d human pose and shape via model-fitting in the loop.
\newblock In \emph{Proceedings of the IEEE/CVF International Conference on Computer Vision}, 2019.

\bibitem[Kolotouros et~al.(2021)Kolotouros, Pavlakos, Jayaraman, and Daniilidis]{kolotouros2021probabilistic}
Nikos Kolotouros, Georgios Pavlakos, Dinesh Jayaraman, and Kostas Daniilidis.
\newblock Probabilistic modeling for human mesh recovery.
\newblock In \emph{Proceedings of the IEEE International Conference on Computer Vision}, pages 11605--11614, 2021.

\bibitem[Li et~al.(2021)Li, Xu, Chen, Bian, Yang, and Lu]{li2021hybrik}
Jiefeng Li, Chao Xu, Zhicun Chen, Siyuan Bian, Lixin Yang, and Cewu Lu.
\newblock Hybrik: A hybrid analytical-neural inverse kinematics solution for 3d human pose and shape estimation.
\newblock In \emph{Proceedings of the IEEE Conference on Computer Vision and Pattern Recognition}, pages 3383--3393, 2021.

\bibitem[Li et~al.(2022)Li, Liu, Zhang, Xu, and Yan]{li2022cliff}
Zhihao Li, Jianzhuang Liu, Zhensong Zhang, Songcen Xu, and Youliang Yan.
\newblock Cliff: Carrying location information in full frames into human pose and shape estimation.
\newblock In \emph{European Conference on Computer Vision}, pages 590--606. Springer, 2022.

\bibitem[Lin et~al.(2021)Lin, Wang, and Liu]{lin2021end-to-end}
Kevin Lin, Lijuan Wang, and Zicheng Liu.
\newblock End-to-end human pose and mesh reconstruction with transformers.
\newblock In \emph{CVPR}, 2021.

\bibitem[Lin et~al.(2024)Lin, Lin, Liang, Liu, and Wang]{lin2024mpt}
Kevin Lin, Chung-Ching Lin, Lin Liang, Zicheng Liu, and Lijuan Wang.
\newblock Mpt: Mesh pre-training with transformers for human pose and mesh reconstruction.
\newblock In \emph{Proceedings of the IEEE/CVF Winter Conference on Applications of Computer Vision}, pages 3415--3425, 2024.

\bibitem[Loper et~al.(2015)Loper, Mahmood, Romero, Pons-Moll, and Black]{SMPL:2015}
Matthew Loper, Naureen Mahmood, Javier Romero, Gerard Pons-Moll, and Michael~J. Black.
\newblock {SMPL}: A skinned multi-person linear model.
\newblock \emph{ACM Transactions on Graphics (Proc. SIGGRAPH Asia)}, 34\penalty0 (6):\penalty0 248:1--248:16, 2015.

\bibitem[Ma et~al.(2023)Ma, Su, Wang, Zhu, and Wang]{ma20233d:VirtualMarkers}
Xiaoxuan Ma, Jiajun Su, Chunyu Wang, Wentao Zhu, and Yizhou Wang.
\newblock 3d human mesh estimation from virtual markers.
\newblock In \emph{Proceedings of the IEEE Conference on Computer Vision and Pattern Recognition}, pages 534--543, 2023.

\bibitem[Mahmood et~al.(2019)Mahmood, Ghorbani, Troje, Pons-Moll, and Black]{AMASS:ICCV:2019}
Naureen Mahmood, Nima Ghorbani, Nikolaus~F. Troje, Gerard Pons-Moll, and Michael~J. Black.
\newblock {AMASS}: Archive of motion capture as surface shapes.
\newblock In \emph{Proceedings of the IEEE/CVF International Conference on Computer Vision}, pages 5442--5451, 2019.

\bibitem[Moon and Lee(2020)]{moon2020i2l}
Gyeongsik Moon and Kyoung~Mu Lee.
\newblock I2l-meshnet: Image-to-lixel prediction network for accurate 3d human pose and mesh estimation from a single rgb image.
\newblock In \emph{European Conference on Computer Vision}, pages 752--768. Springer, 2020.

\bibitem[Mugaludi et~al.(2021)Mugaludi, Kundu, Jampani, et~al.]{mugaludi2021aligning}
Ramesha~Rakesh Mugaludi, Jogendra~Nath Kundu, Varun Jampani, et~al.
\newblock Aligning silhouette topology for self-adaptive 3d human pose recovery.
\newblock \emph{Advances in Neural Information Processing Systems}, 34:\penalty0 4582--4593, 2021.

\bibitem[Paszke et~al.(2019)Paszke, Gross, Massa, Lerer, Bradbury, Chanan, Killeen, Lin, Gimelshein, Antiga, et~al.]{paszke2019pytorch}
Adam Paszke, Sam Gross, Francisco Massa, Adam Lerer, James Bradbury, Gregory Chanan, Trevor Killeen, Zeming Lin, Natalia Gimelshein, Luca Antiga, et~al.
\newblock Pytorch: An imperative style, high-performance deep learning library.
\newblock \emph{Advances in Neural Information Processing Systems}, 32, 2019.

\bibitem[Pavlakos et~al.(2018)Pavlakos, Zhu, Zhou, and Daniilidis]{pavlakos2018learning}
Georgios Pavlakos, Luyang Zhu, Xiaowei Zhou, and Kostas Daniilidis.
\newblock Learning to estimate 3d human pose and shape from a single color image.
\newblock In \emph{Proceedings of the IEEE Conference on Computer Vision and Pattern Recognition}, pages 459--468, 2018.

\bibitem[Petrovich et~al.(2021)Petrovich, Black, and Varol]{petrovich2021action}
Mathis Petrovich, Michael~J Black, and G{\"u}l Varol.
\newblock Action-conditioned 3d human motion synthesis with transformer vae.
\newblock In \emph{Proceedings of the IEEE/CVF International Conference on Computer Vision}, pages 10985--10995, 2021.

\bibitem[Rempe et~al.(2021)Rempe, Birdal, Hertzmann, Yang, Sridhar, and Guibas]{rempe2021humor}
Davis Rempe, Tolga Birdal, Aaron Hertzmann, Jimei Yang, Srinath Sridhar, and Leonidas~J Guibas.
\newblock Humor: 3d human motion model for robust pose estimation.
\newblock In \emph{Proceedings of the IEEE International Conference on Computer Vision}, pages 11488--11499, 2021.

\bibitem[Rodrigues et~al.(2022)Rodrigues, Antunes, Seewald, Bazo, dos Reis, dos Santos, Righi, Junior, da~Costa, Bertollo, et~al.]{rodrigues2022multi}
Vinicius~F Rodrigues, Rodolfo~S Antunes, Lucas~A Seewald, Rodrigo Bazo, Eduardo~S dos Reis, Uelison~JL dos Santos, Rodrigo da~R Righi, Luiz G da~S Junior, Cristiano~A da Costa, Felipe~L Bertollo, et~al.
\newblock A multi-sensor architecture combining human pose estimation and real-time location systems for workflow monitoring on hybrid operating suites.
\newblock \emph{Future Generation Computer Systems}, 135:\penalty0 283--298, 2022.

\bibitem[Sengupta et~al.(2020)Sengupta, Budvytis, and Cipolla]{sengupta2020synthetic:STRAP}
Akash Sengupta, Ignas Budvytis, and Roberto Cipolla.
\newblock Synthetic training for accurate 3d human pose and shape estimation in the wild.
\newblock \emph{arXiv preprint arXiv:2009.10013}, 2020.

\bibitem[Sengupta et~al.(2021{\natexlab{a}})Sengupta, Budvytis, and Cipolla]{sengupta2021hierarchical:STRAPV3}
Akash Sengupta, Ignas Budvytis, and Roberto Cipolla.
\newblock Hierarchical kinematic probability distributions for 3d human shape and pose estimation from images in the wild.
\newblock In \emph{Proceedings of the IEEE International Conference on Computer Vision}, pages 11219--11229, 2021{\natexlab{a}}.

\bibitem[Sengupta et~al.(2021{\natexlab{b}})Sengupta, Budvytis, and Cipolla]{sengupta2021probabilistic:STRAPV2}
Akash Sengupta, Ignas Budvytis, and Roberto Cipolla.
\newblock Probabilistic 3d human shape and pose estimation from multiple unconstrained images in the wild.
\newblock In \emph{Proceedings of the IEEE Conference on Computer Vision and Pattern Recognition}, pages 16094--16104, 2021{\natexlab{b}}.

\bibitem[Shen et~al.(2018)Shen, Qu, Zhang, and Yu]{shen2018wasserstein:WDGRL}
Jian Shen, Yanru Qu, Weinan Zhang, and Yong Yu.
\newblock Wasserstein distance guided representation learning for domain adaptation.
\newblock In \emph{Proceedings of the AAAI Conference on Artificial Intelligence}, 2018.

\bibitem[Song et~al.(2020)Song, Chen, and Hilliges]{song2020human:LGD}
Jie Song, Xu Chen, and Otmar Hilliges.
\newblock Human body model fitting by learned gradient descent.
\newblock In \emph{European Conference on Computer Vision}, pages 744--760. Springer, 2020.

\bibitem[Tripathi et~al.(2020)Tripathi, Ranade, Tyagi, and Agrawal]{tripathi2020posenet3d}
Shashank Tripathi, Siddhant Ranade, Ambrish Tyagi, and Amit Agrawal.
\newblock Posenet3d: Learning temporally consistent 3d human pose via knowledge distillation.
\newblock In \emph{International Conference on 3D Vision}, pages 311--321. IEEE, 2020.

\bibitem[Tripathi et~al.(2023)Tripathi, M{\"u}ller, Huang, Taheri, Black, and Tzionas]{tripathi20233d}
Shashank Tripathi, Lea M{\"u}ller, Chun-Hao~P Huang, Omid Taheri, Michael~J Black, and Dimitrios Tzionas.
\newblock 3d human pose estimation via intuitive physics.
\newblock In \emph{Proceedings of the IEEE Conference on Computer Vision and Pattern Recognition}, pages 4713--4725, 2023.

\bibitem[Tung et~al.(2017)Tung, Tung, Yumer, and Fragkiadaki]{tung2017self}
Hsiao-Yu Tung, Hsiao-Wei Tung, Ersin Yumer, and Katerina Fragkiadaki.
\newblock Self-supervised learning of motion capture.
\newblock \emph{Advances in Neural Information Processing Systems}, 30, 2017.

\bibitem[Tzeng et~al.(2017)Tzeng, Hoffman, Saenko, and Darrell]{tzeng2017adversarial:ADDA}
Eric Tzeng, Judy Hoffman, Kate Saenko, and Trevor Darrell.
\newblock Adversarial discriminative domain adaptation.
\newblock In \emph{Proceedings of the IEEE Conference on Computer Vision and Pattern Recognition}, pages 7167--7176, 2017.

\bibitem[Von~Marcard et~al.(2018)Von~Marcard, Henschel, Black, Rosenhahn, and Pons-Moll]{von2018recovering:3dpw}
Timo Von~Marcard, Roberto Henschel, Michael~J Black, Bodo Rosenhahn, and Gerard Pons-Moll.
\newblock Recovering accurate 3d human pose in the wild using imus and a moving camera.
\newblock In \emph{European Conference on Computer Vision}, pages 601--617, 2018.

\bibitem[Weng et~al.(2019)Weng, Curless, and Kemelmacher-Shlizerman]{weng2019photo}
Chung-Yi Weng, Brian Curless, and Ira Kemelmacher-Shlizerman.
\newblock Photo wake-up: 3d character animation from a single photo.
\newblock In \emph{Proceedings of the IEEE Conference on Computer Vision and Pattern Recognition}, pages 5908--5917, 2019.

\bibitem[Xu et~al.(2022)Xu, Zhang, Zhang, and Tao]{xu2022vitpose}
Yufei Xu, Jing Zhang, Qiming Zhang, and Dacheng Tao.
\newblock Vi{TP}ose: Simple vision transformer baselines for human pose estimation.
\newblock In \emph{Advances in Neural Information Processing Systems}, 2022.

\bibitem[Yu et~al.(2021)Yu, Wang, Xu, Ni, Zhao, Wang, and Zhang]{yu2021skeleton2mesh}
Zhenbo Yu, Junjie Wang, Jingwei Xu, Bingbing Ni, Chenglong Zhao, Minsi Wang, and Wenjun Zhang.
\newblock Skeleton2mesh: Kinematics prior injected unsupervised human mesh recovery.
\newblock In \emph{Proceedings of the IEEE International Conference on Computer Vision}, pages 8619--8629, 2021.

\bibitem[Zanfir et~al.(2020)Zanfir, Bazavan, Xu, Freeman, Sukthankar, and Sminchisescu]{zanfir2020weakly}
Andrei Zanfir, Eduard~Gabriel Bazavan, Hongyi Xu, William~T Freeman, Rahul Sukthankar, and Cristian Sminchisescu.
\newblock Weakly supervised 3d human pose and shape reconstruction with normalizing flows.
\newblock In \emph{European Conference on Computer Vision}, pages 465--481. Springer, 2020.

\bibitem[Zanfir et~al.(2021)Zanfir, Bazavan, Zanfir, Freeman, Sukthankar, and Sminchisescu]{zanfir2021neural}
Andrei Zanfir, Eduard~Gabriel Bazavan, Mihai Zanfir, William~T Freeman, Rahul Sukthankar, and Cristian Sminchisescu.
\newblock Neural descent for visual 3d human pose and shape.
\newblock In \emph{Proceedings of the IEEE Conference on Computer Vision and Pattern Recognition}, pages 14484--14493, 2021.

\bibitem[Zhang et~al.(2021{\natexlab{a}})Zhang, Tian, Zhou, Ouyang, Liu, Wang, and Sun]{pymaf2021}
Hongwen Zhang, Yating Tian, Xinchi Zhou, Wanli Ouyang, Yebin Liu, Limin Wang, and Zhenan Sun.
\newblock Pymaf: 3d human pose and shape regression with pyramidal mesh alignment feedback loop.
\newblock In \emph{Proceedings of the IEEE International Conference on Computer Vision}, 2021{\natexlab{a}}.

\bibitem[Zhang et~al.(2021{\natexlab{b}})Zhang, Zhang, Bogo, Pollefeys, and Tang]{zhang2021learning}
Siwei Zhang, Yan Zhang, Federica Bogo, Marc Pollefeys, and Siyu Tang.
\newblock Learning motion priors for 4d human body capture in 3d scenes.
\newblock In \emph{Proceedings of the IEEE International Conference on Computer Vision}, pages 11343--11353, 2021{\natexlab{b}}.

\bibitem[Zhou et~al.(2019)Zhou, Barnes, Lu, Yang, and Li]{zhou2019continuity:6DRot}
Yi Zhou, Connelly Barnes, Jingwan Lu, Jimei Yang, and Hao Li.
\newblock On the continuity of rotation representations in neural networks.
\newblock In \emph{Proceedings of the IEEE Conference on Computer Vision and Pattern Recognition}, pages 5745--5753, 2019.

\end{thebibliography}
}
\clearpage
\setcounter{page}{1}
\maketitlesupplementary

\section{Qualitative Results}

\begin{figure}
\centering
\includegraphics[width=0.47\textwidth]{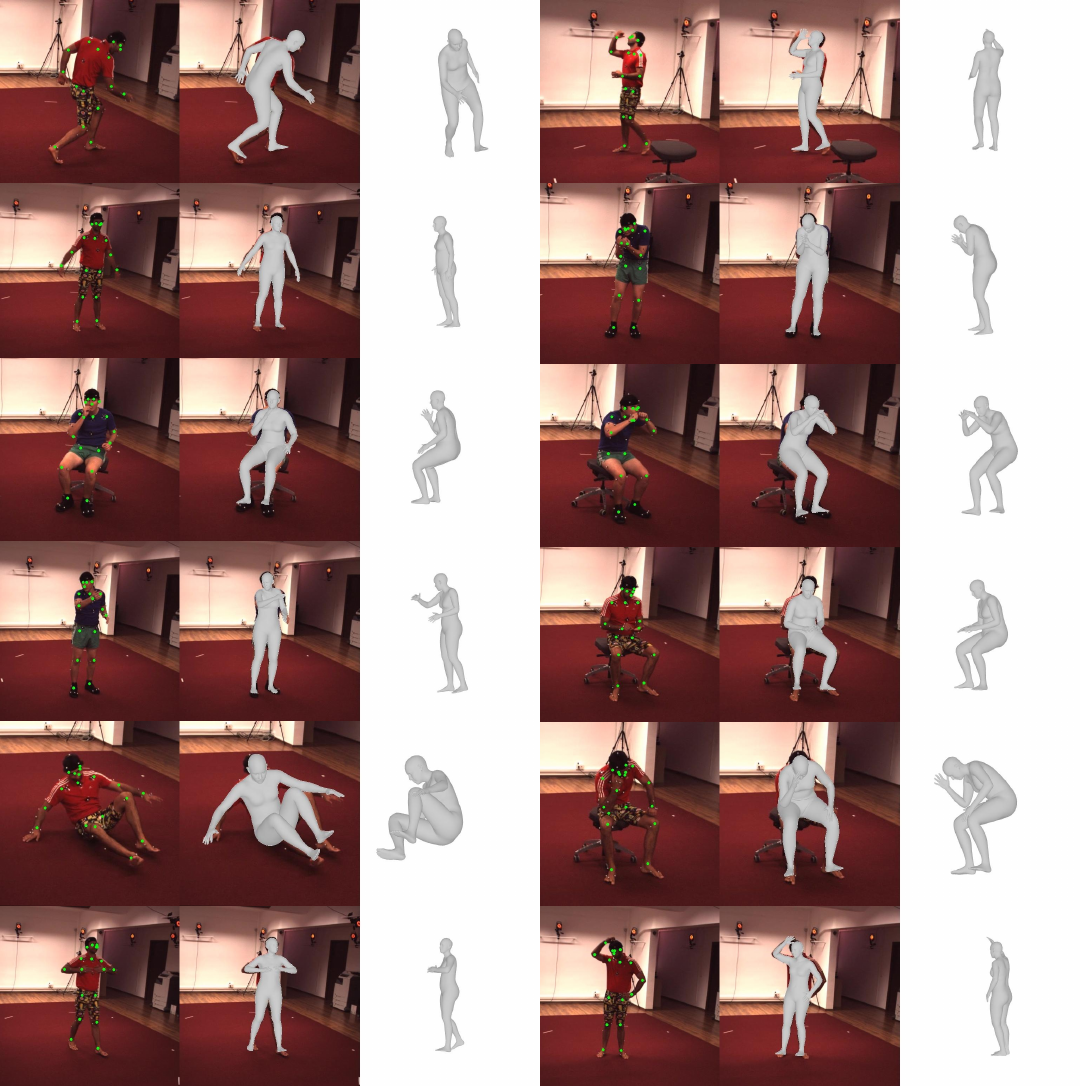}
\caption{\textbf{Qualitative results on the H3.6M dataset \cite{h36m_pami}.} For each sample, the first column displays the input image and 2D keypoint detections, the second column shows the SMPL mesh overlaid on the image, and the third column presents the SMPL mesh from the side view. Domain-adapted model is used to generate these qualitative results.}
\label{fig:h36m-qual}
\end{figure}

We showcase qualitative results on the H3.6M \cite{h36m_pami} dataset in \cref{fig:h36m-qual}. Moreover, we provide qualitative video results on the project website: \url{https://key2mesh.github.io/}. Our approach involves applying the Key2Mesh (adapted) model to each individual video frame, with predictions presented without any smoothing. As there are no 2D keypoints on hands and feet in the OpenPose detections, we occasionally observe jittery artifacts on the extremities. Incorporating additional 2D keypoints on the extremities could be addressed in future work.

\section{Further Implementation Details}

\begin{figure}
\centering
\includegraphics[width=0.47\textwidth]{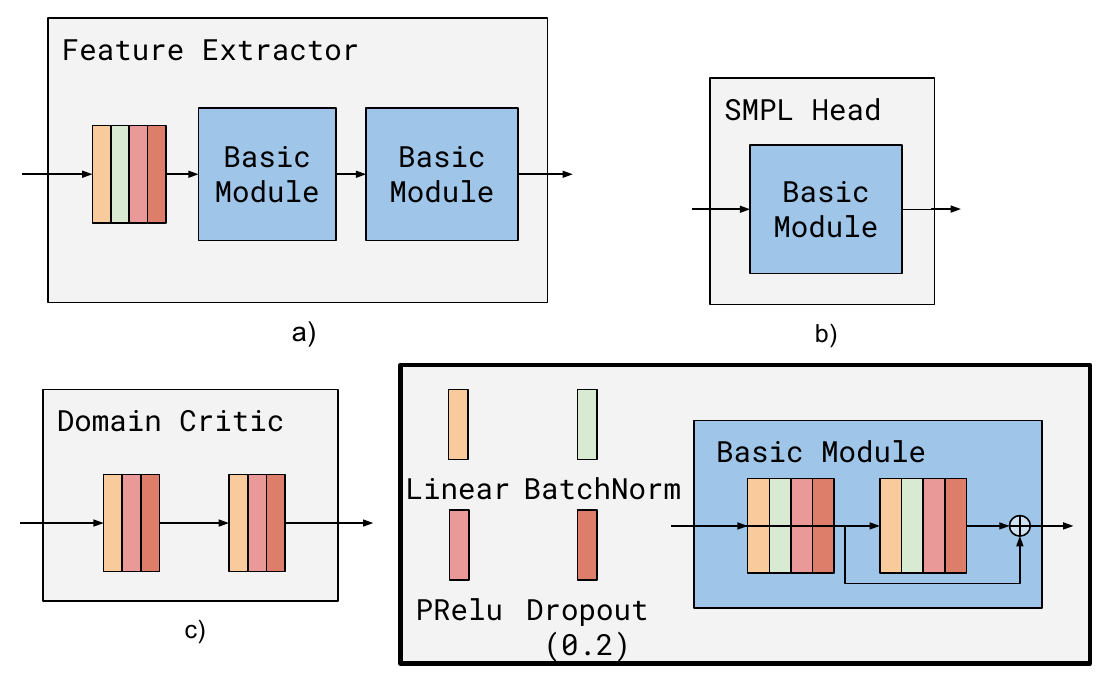}
\caption{\textbf{Overview of Key2Mesh model architecture.}}
\label{fig:supp-model-arch}
\end{figure}

\noindent \textbf{Model Architecture Details.}
\cref{fig:supp-model-arch} illustrates the architecture of our model, consisting of three key components: the Feature Extractor, the SMPL Head, and the Domain Critic. For each component, we adopted a standard multi-layer perceptron (MLP) architecture, incorporating skip connections as shown in the figure. The linear layers were equipped with 1024 neurons, and we introduced batch normalization after each linear layer, except in the case of the Domain Critic. In the early stages of our experiments, we observed that excluding the normalization layer inside the Domain Critic led to better performance. We utilize parametric Rectified Linear Unit (pReLU) and Dropout layers, setting the dropout probability to 0.2.

\noindent \textbf{Model Selection.}
While adapting to the test dataset, selecting a model becomes less straightforward due to the absence of 3D labels. As a result, benchmark metrics like PA-MPJPE, MPJPE, and PVE cannot be calculated during evaluation and model selection. To obtain the models adapted to the test sets of H3.6M and 3DPW, as depicted in the final rows of Tab. 1 and Tab. 2, we employ the training subjects from the H3.6M dataset and the validation split from 3DPW during the model selection process. We calculate PA-MPJPE every 500 training steps on these sets and checkpoint the model that achieves the best PA-MPJPE during this process.

\section{Training Data Sampling from Unpaired 3D Human Body Data}
In pre-training, we sample a 3D human body encoded with SMPL \cite{SMPL:2015} from the MoCap domain and apply a range of augmentations. For a given SMPL sample, we follow LGD's \cite{song2020human:LGD} augmentation pipeline and apply random global rotations to simulate different views by using yaw angles drawn uniformly from the range of $-180^\circ$ to $+180^\circ$, and roll and pitch angles sampled uniformly within $-20^\circ$ to $20^\circ$. We also randomly occlude body joints with $20\%$ probability. In addition to LGD's augmentation pipeline, we try to simulate the jitter introduced by the pose estimator on the visual data for each keypoint ($x$). To achieve this, we perturb keypoints randomly: $x = x + \epsilon$, where $\epsilon \sim \mathcal{N}(0, I)$.

\section{Using Different 2D Pose Estimators}

\begin{table}[H]
\centering
\begin{tabular}{|c|c|c|}
\hline
2D Pose Estimator & PA-MPJPE$\downarrow$ & MPJPE$\downarrow$\\ \hline \hline
OpenPose \cite{cao2017realtime:OpenPose} & 51.4 & 108.1 \\ \hline
ViTPose \cite{xu2022vitpose} & \textbf{49.3} & \textbf{104.2} \\ \hline
\end{tabular}

\caption{\textbf{Comparing the use of OpenPose \cite{cao2017realtime:OpenPose} and ViTPose \cite{xu2022vitpose} as 2D pose estimators in the pipeline.} We report PA-MPJPE and MPJPE (both in mm) based on the H3.6M evaluation subjects after applying our domain adaptation stage using H3.6M training subjects.}
\label{table:pose-estimator-effect}
\end{table}

In table \cref{table:pose-estimator-effect}, we compare the performance of using OpenPose \cite{cao2017realtime:OpenPose} and ViTPose \cite{xu2022vitpose} as 2D pose estimators in the pipeline. We observe that utilizing ViTPose results in lower PA and MPJPE scores compared to OpenPose. This underscores the potential for incorporating different pose estimators into Key2Mesh's pipeline, with a better-performing pose estimator leading to improved SMPL estimation accuracy.

\section{Impact of the Domain Adaptation on the Features}

\begin{figure}[H]
\centering
\includegraphics[width=0.47\textwidth]{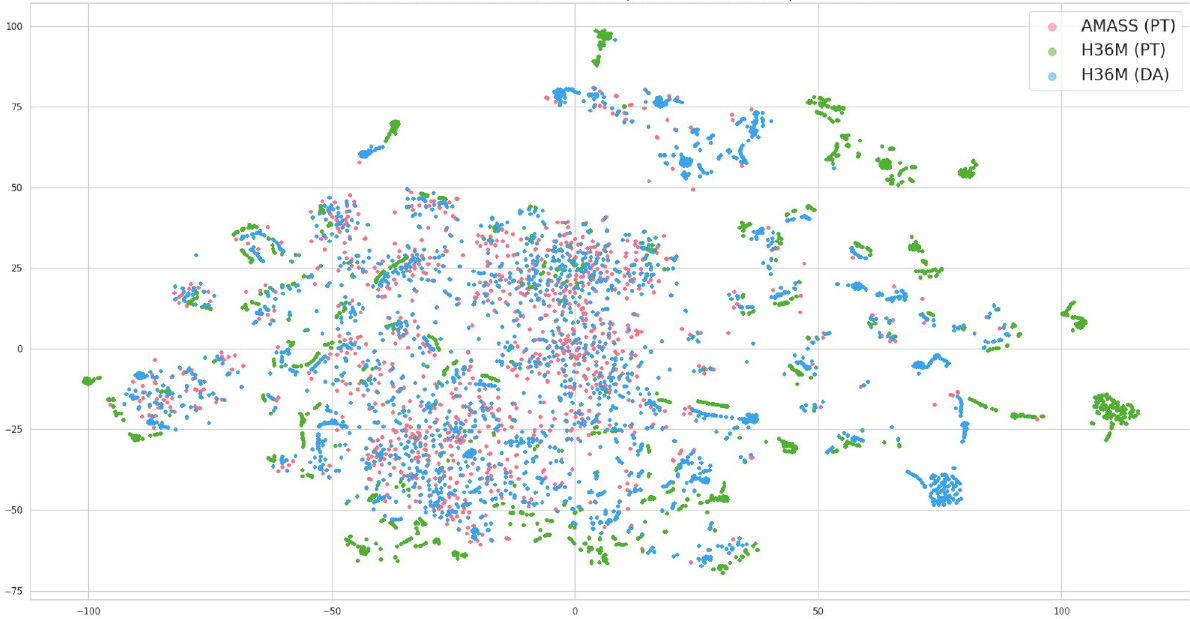}
\caption{\textbf{t-SNE visualization of features extracted by pre-trained and domain-adapted feature extractors on H3.6M.} Best viewed when zoomed in.}
\label{fig:tsne-features}
\end{figure}

The domain-adapted feature extractor is constrained to produce features that mimic those of the source domain when the domain of the input 2D poses changes. As illustrated in \cref{fig:tsne-features}, the t-SNE plot shows that features generated by the domain-adapted feature extractor on the H3.6M dataset (blue) exhibit improved alignment with the source domain, as represented by features generated by the pre-trained feature extractor on the AMASS dataset (pink), compared to features produced solely by the pre-trained feature extractor (green). This aligns with our intuition regarding domain adaptation, as it improves the performance of SMPL-Head when transitioning between domains by operating on features that more closely resemble those from its training set.

\section{Failure Scenarios}

\begin{figure}[H]
\centering
\includegraphics[width=0.47\textwidth]{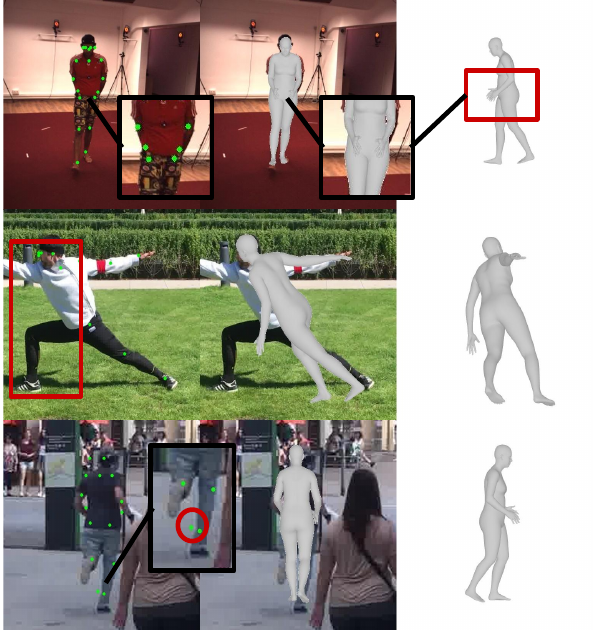}
\caption{\textbf{Some failure cases.}
In the first row, the algorithm encounters challenges related to depth ambiguity, leading to inaccurate hand position detection. In the second and third rows, Key2Mesh exhibits issues with missing or incorrect keypoint detection. The model's exclusive reliance on 2D keypoints renders it susceptible to errors in cases where keypoints are either absent or detected erroneously.
}
\label{fig:failure-cases}
\end{figure}

\cref{fig:failure-cases} illustrates various cases that highlight the limitations of Key2Mesh. 

\section{Limitations and Future work} While our pipeline shows promising performance in predicting 3D human pose and shape from 2D keypoints, inherent ambiguity in representing 3D pose and shape through 2D keypoints poses fundamental challenges. Future work might involve integrating temporal information to enhance performance.

\end{document}